\documentclass[letterpaper, 10 pt, conference]{ieeeconf}  % Comment this line out if you need a4paper

\IEEEoverridecommandlockouts                              % This command is only needed if 
\usepackage{graphics} % for pdf, bitmapped graphics files
\usepackage{amsmath} % assumes amsmath package installed
\usepackage{amssymb}  % assumes amsmath package installed
\usepackage{todonotes}
\usepackage{booktabs}
\usepackage{multirow}
\usepackage{url}

\title{\LARGE \bf
Neural Network Virtual Sensors for Fuel Injection Quantities with Provable Performance Specifications
}

\author{Eric Wong$^{1}$, Tim Schneider$^{2}$, Joerg Schmitt$^{3}$, Frank R. Schmidt,$^{4}$ J. Zico Kolter$^{5}$% <-this % stops a space
\thanks{$^{1}$Machine Learning Department, Carnegie Mellon University,
Pittsburgh, PA 15213, USA {\tt\small ericwong@cs.cmu.edu}}%
\thanks{$^{2}$Robert Bosch GmbH and Bosch Center for Artificial Intelligence, Germany
        {\tt\small Fixed-term.tim.schneider3@de.bosch.com}}%
\thanks{$^{3}$Robert Bosch GmbH, Stuttgart, Germany\newline
        {\tt\small Joerg.Schmitt2@de.bosch.com}}%
\thanks{$^{4}$Bosch Center for Artificial Intelligence, Renningen, Germany
        {\tt\small frank.r.schmidt@de.bosch.com}}%
\thanks{$^{5}$Computer Science Department, Carnegie Mellon University and Bosch Center for Artificial Intelligence, Pittsburgh, PA 15213, USA
        {\tt\small zkolter@cs.cmu.edu}}%
}

\begin{document}

\maketitle
\thispagestyle{empty}
\pagestyle{empty}

%%%%%%%%%%%%%%%%%%%%%%%%%%%%%%%%%%%%%%%%%%%%%%%%%%%%%%%%%%%%%%%%%%%%%%%%%%%%%%%%
\begin{abstract}

Recent work has shown that it is possible to learn neural networks with provable guarantees on the output of the model when subject to input perturbations, however these works have focused primarily on defending against adversarial examples for image classifiers. In this paper, we study how these provable guarantees can be naturally applied to other real world settings, namely getting performance specifications for robust virtual sensors measuring fuel injection quantities within an engine. We first demonstrate that, in this setting, even simple neural network models are highly susceptible to reasonable levels of adversarial sensor noise, which are capable of increasing the mean relative error of a standard neural network from 6.6\% to 43.8\%. We then leverage methods for learning provably robust networks and verifying robustness properties, resulting in a robust model which we can provably guarantee has at most 16.5\% mean relative error under any sensor noise. Additionally, we show how specific intervals of fuel injection quantities can be targeted to maximize robustness for certain ranges, allowing us to train a virtual sensor for fuel injection which is provably guaranteed to have at most 10.69\% relative error under noise while maintaining 3\% relative error on non-adversarial data within normalized fuel injection ranges of 0.6 to 1.0. 

\end{abstract}

%%%%%%%%%%%%%%%%%%%%%%%%%%%%%%%%%%%%%%%%%%%%%%%%%%%%%%%%%%%%%%%%%%%%%%%%%%%%%%%%
\section{INTRODUCTION}

Although neural networks have made remarkable progress in achieving state-of-the-art performance in vision and language tasks over recent years, other domains have been comparatively slower at incorporating these types of models into their systems. For higher-states applications such as those in driving, security, or health care systems, the lack of performance guarantees and the inability to interpret the model constitute critical shortcomings that prevent neural networks from being more widely adopted. The unpredictability and brittleness of neural networks is exemplified by a now well-studied phenomenon called \emph{adversarial examples} \cite{szegedy2014intriguing}, where datapoints that look indistinguishable from ``normal'' examples are specifically perturbed to be misclassified by machine learning systems. These adversarial examples originally focused on $\ell_\infty$-bounded perturbations, which model a small amount of noise added to each pixel in the vision setting \cite{goodfellow2015explaining}, but have since expanded to other perturbations ranging from imperceptible perturbations such as $\ell_p$-bounded additive noise \cite{sharif2018suitability, tramer2019adversarial, maini2019adversarial}, image transformations such as rotations or translations \cite{engstrom2017rotation, xiao2018spatially}, and Wasserstein perturbations \cite{wong2019wasserstein, levine2019wasserstein}, to more visible, real-world perturbations such as adversarial stickers \cite{li2019adversarial, brown2017adversarial}, clothing \cite{wu2019making}, glasses \cite{sharif2017adversarial}, and textures \cite{pmlr-v80-athalye18b}. 

As a result, a natural prerequisite towards deploying neural networks in higher-stakes applications is to learn a model which performs well not just under the typical setting of test-set performance, but to also be robust to perturbed inputs. To this end, there has been a great deal of work towards training networks which are not only empirically robust \cite{madry2018towards, zhang2019theoretically}, but also provably robust \cite{wong2018provable, wong2018scaling, raghunathan2018semi, mirman2018diff, gowal2018interval}, the latter of which is able to give provable guarantees on the performance of the neural network against \emph{any} perturbation within a given set. Although these methods could potentially be used to verify crucial safety and reliability requirements in other domains, provably robust training methods have largely been focused on vision and language domains \cite{jia2019certified}, where they are largely limited to medium-scale networks and haven't seen much use outside the context of defending against adversarial examples. 

In this paper, we study how provably robust networks can be leveraged for real-world problems beyond adversarial examples in vision and language problems. Specifically, we focus on the problem of learning a virtual sensor for use in an engine controller, where the goal is to learn a model which accurately estimates the amount of fuel which was injected into the engine of a vehicle based on various sensors readings. Although a virtual sensor for fuel injection may never be subject to worst-case adversarial examples in the real world, reliably and robustly predicting accurate fuel injection quantities under noisy conditions has direct consequences on fuel efficiency and engine safety. Consequently, provable guarantees against adversarial examples can be leveraged not as a defense mechanism against adversarial examples, but instead as a way to learn model specifications which guarantee certain levels of performance of the neural network under entire sets of noise. 

We first demonstrate that even simple neural network models suffer from adversarial examples in this setting, where perturbed sensor readings can drastically degrade the model performance from 6.6\% to 43.8\% mean relative error, subject to sensor perturbations of 1\% for time series data from one sensor and 0.1\% for data from a second sensor. To improve the robustness of the neural network and obtain performance guarantees, we adapt provably robust training methods based on duality \cite{wong2018provable, wong2018scaling} to the regression setting, and train a provably robust model which achieves only 16.4\% mean relative error when subject to adversarial sensor perturbations. More importantly, the provably robust and verification methods developed for adversarial examples produce bounds which are specifications on the worst-case performance of the model: our virtual sensor is provably guaranteed to have at most 16.5\% mean relative error under these noise levels. Finally, in some scenarios, it may be desirable to have a model which has better performance specifications within a limited output range instead, rather than having overall robustness across the entire output range. We show how the robustness of the virtual sensor can be tuned to target a limited set of output ranges, resulting in a model which achieves 3\% relative test error and is provably guaranteed to have at most 10.69\% worst-case relative error when predicting normalized fuel injection quantities 0.6 and 1.0 under any sensor noise. Notably, this is comparable in performance to that of non-robust standard training, which achieves 2.7\% relative error within the same output range but with a worse guarantee of 17.78\% worst-case relative error. 

\section{BACKGROUND}
After their initial discovery \cite{szegedy2014intriguing}, adversarial examples were initially generated by using a single gradient step called the fast gradient sign method (FGSM) \cite{goodfellow2015explaining}, which was later improved by adding randomized initialization \cite{tramer2017ensemble}, taking multiple smaller steps \cite{kurakin2017adversarial}, and using momentum \cite{dong2018boosting}. By training on the adversarially perturbed examples instead of the original examples, it is possible to learn models which are empirically robust to an adversarial attack, also known as adversarial training. Training against a projected gradient descent adversary is recognized as one of the earliest known defenses against adversarial examples that remains robust to this day \cite{madry2018towards}, and has been further improved in subsequent work to achieve higher performance \cite{zhang2019theoretically}, generalize to different threat models \cite{tramer2019adversarial, maini2019adversarial}, and combine with other heuristic defenses \cite{mosbach2018logit, yang2019me}. Variations of adversarial training were proposed to speed up the process, such as free adversarial training \cite{shafahi2019adversarial}, reducing the complexity of computing gradients \cite{zhang2019you}, and a return to single step FGSM adversarial training \cite{wong2019fast}. 

Other methods beyond adversarial training for mitigating the effect of adversarial examples include preprocessing techniques \cite{guo2017countering, buckman2018thermometer, song2017pixeldefend} and detection algorithms \cite{metzen2017detecting, feinman2017detecting}. However, a significant number of defenses against adversarial examples  \cite{papernot2016distillation, lu2017no, kannan2018adversarial, tao2018attacks} were ultimately shown to be ineffective \cite{carlini2016defensive, carlini2017towards, athalye2017synthesizing, engstrom2018evaluating, carlini2019ami}, and several papers began defeating adversarial defenses en masse \cite{uesato2018adversarial, athalye2018obfuscated} as well as recommending best practices to ensure a proper evaluation for adversarial defenses \cite{carlini2019evaluating}. 

Some work has looked into designing unbreakable defenses known as provable or certified defenses against adversarial attacks for neural networks. In contrast to other approaches, these methods calculate a guaranteed bound on the output of the network over some region of the input space, typically taken to be an $\ell_p$ ball around a data point. Some of these bounds can be computationally expensive and can only be used to verify trained networks, relying on satisfiability modulo theory (SMT) solvers \cite{katz2017reluplex, huang2017safety, ehlers2017formal}, mixed integer linear programming (MILP) \cite{tjeng2017evaluating}, semidefinite programming \cite{raghunathan2018semidefinite}, and linear programming \cite{wong2018provable, salman2019convex}. Other bounds which are looser but more efficiently computable can actually be optimized during training, utilizing techniques for distributional robustness \cite{sinha2017certifying}, duality \cite{wong2018provable, wong2018scaling}, more efficient semidefinite programming formulations \cite{raghunathan2018certified}, abstract interpretations \cite{gehr2018ai2, mirman2018differentiable}, interval bound propagation \cite{gowal2018effectiveness}, and randomized smoothing \cite{cohen2019certified, salman2019provably}. Other work has designed theoretically motivated training heuristics which can be independently verified at test time as being robust \cite{xiao2018training, croce2018provable}.  

Beyond adversarial examples, other relevant work includes those studying model verification in contexts beyond image classification, typically within the context of controller verification. To name a few, hybrid controllers for automated highway systems were verified to be safe by design with game theoretic techniques \cite{lygeros1998verified}, and frameworks for verifying controller software for manufacturing plants were developed for programmable logic controllers \cite{gerber2010complete}. While some work has looked at verifying properties of neural networks for safety critical applications, they are mostly limited to measuring the confidence and monitoring the performance of an existing neural network without offering any formal guarantees \cite{schumann2003verification, gupta2004tool}. Most relevant to our work are those which give provable guarantees for neural network models, such as using bounded model checking techniques to verify safety properties of a controller for the classic Cart Pole System (inverted pendulum) with an SMT solver \cite{scheibler2015towards}, as well as recent work which verified the safety of an autonomous robot controller using satisfiability modulo convex optimization \cite{sun2019formal}. 

\section{ROBUST FUEL INJECTION}
\label{section:robustregression}
Unlike past work which has largely focused on large-scale image classification, in this paper we study the regression problem of learning a neural network virtual sensor for fuel injection. The virtual sensor system predicts the amount of injected fuel during a specific time interval, based on two sensors. The first sensor provides a time series of sensor data while the second sensor is essentially constant within the time interval. More accurate fuel injection predictions result in better fuel efficiency and safety considerations, however the sensor readings are subject to measurement noise. The goal is to not only train a neural network which accurately predicts fuel injection quantities while being robust to measurement noise from the sensors, but to also generate certificates which can guarantee the worst-case performance of the model under noise and serve as a certified specification for the model. 

To train a neural network robust to noise, we leverage duality-based methods for learning provably robust networks \cite{wong2018provable, wong2018scaling}, which converts inputs, layers, and the loss into their dual components to construct a bound on the output of the network. Specifically, for a given neural network $f$ and input $x$, these methods are able to compute a bound $J(x)$ on the worst case output of the network subject to some perturbation set $\mathcal{B}(x)$ around $x$ 
\begin{equation}
\max_{z \in \mathcal B(x)} f(z)\cdot c \leq J(x; c)
\end{equation}
for any constant $c$ (we use a network with one output unit for simplicity as that is sufficient for our regression setting, but the original approach from \cite{wong2018scaling} applies to multiple outputs). By choosing $c = \{-1, 1\}$, one can compute upper and lower bounds on the output of a neural network.\footnote{These bounds are computed with the open-source implementation of the duality based approach available at \url{https://github.com/locuslab/convex_adversarial}} 

To adapt the method to the regression setting, we can use the duality-based method to directly compute lower and upper bounds on the output to the network and use these to compute a bound on the mean squared error. Specifically, let 
\begin{equation}
\ell \leq \min_{z \in \mathcal B(x)} f(z), \quad \max_{z \in \mathcal B(x)} f(z) \leq u
\end{equation} 
be lower and upper bounds $(\ell,u)$ on the output of a network $f$ subject to perturbations in $\mathcal B(x)$ around an example $x$. Then, we can bound the worst-case mean squared error within the same set $\mathcal B(x)$ with respect to the output $y$ with 
\begin{equation}
\max_{z \in \mathcal B(x)} (f(z)-y)^2 \leq \max((\ell-y)^2, (u-y)^2)
\end{equation}
which we will call \emph{robust mean squared error}. Training provably robust regression models then amounts to running backpropagation on the robust mean squared error, and the lower and upper bounds serve as guaranteed certificates on the output of the neural network. 
%Due to its modular nature, we can adapt the method to the regression setting by replacing the loss from the classification setting (difference between classes) to the regression setting (squared error) and computing the corresponding dual loss. This is summarized in Lemma \ref{lem:regression}. 

%\begin{lemma}
%\label{lem:regression}
%Let the loss for a neural network $f$ on a data point $x$ with label $y$ be $\ell(f(x),y) = \frac{1}{2}(f(x)-y)^2)$, or mean squared error. Then, the corresponding dual of the loss, $\ell^\star(\nu)$ is 
%\begin{equation}
%\ell^\star(\nu) = -\frac{1}{2}\nu^2 + y\nu 
%%\min_z \frac{1}{2}(z-y)^2 + z\nu = -\frac{1}{2}\nu^2 + y\nu 
%\end{equation}
%\end{lemma}
%
%\begin{proof}
%Following the proof from \cite{wong2018scaling}, note that the dual loss is computed by solving 
%\begin{equation}
%\ell^\star(\nu) = \min_z \ell(z,y) + z\nu
%\end{equation}
%For the mean squared error loss, the minimum occurs at $z=y-\nu$. Plugging this results in the dual of the mean squared error loss. 
%\end{proof}
%
%With this lemma, we can replace the classification loss from \cite{wong2018scaling} with the regression loss, and compute guaranteed bounds on the mean squared error of neural networks with ReLU non-linearities over bounded perturbation regions. By backpropagating to minimize the bound on mean squared error, we can now learn a fuel injection controller with non-vacuous provable guarantees. 

\begin{table*}[t]
\vspace{0.3em}
\caption{Fuel injection performance over various fuel injection ranges and training methods}
\vspace{-1.3em}
\label{table:performance}
\begin{center}
\begin{tabular}{clrrrrrr}
\toprule
&                          &    [0.0-0.2)  &  [0.2-0.4)  & [0.4-0.6)  &  [0.6-0.8) & [0.8-1.0] & [0.0-1.0]\\
\midrule
\multirow{5}*{Standard training}
& Relative error           &   15.96\%  &    8.17\%  &    4.52\%  &    2.87\%  &    2.53\% &  6.61\%\\
& Noise error              &   22.89\%  &    9.94\%  &    5.70\%  &    3.77\%  &    3.13\% &  8.76\%\\
& PGD error                &  119.03\%  &   46.94\%  &   28.15\%  &   19.39\%  &   14.74\% & {\bf 43.88\%}\\
& MILP bound               &  125.43\%  &   48.39\%  &   28.89\%  &   19.82\%  &   15.14\% & 45.63\%\\
& Dual bound               &  139.48\%  &   52.96\%  &   31.21\%  &   21.31\%  &   16.25\% & 50.09\%\\
\midrule 
\multirow{5}*{Noise data augmentation}
& Relative error           &   15.63\%  &    8.64\%  &    5.00\%  &    2.93\%  &    2.61\% & 6.79\% \\
& Noise error              &   18.89\%  &    9.47\%  &    5.60\%  &    3.50\%  &    2.97\% & 7.86\% \\
& PGD error                &   85.26\%  &   35.92\%  &   23.05\%  &   16.04\%  &   12.56\% & \bf{33.38\%} \\
& MILP bound               &   91.05\%  &   37.36\%  &   23.98\%  &   16.53\%  &   12.90\% & 35.07\% \\
& Dual bound               &  109.61\%  &   41.90\%  &   26.03\%  &   18.06\%  &   14.19\% & 40.29\% \\
\midrule
\multirow{5}*{Robust training}
& Relative error           &   19.79\%  &   11.45\%  &    9.50\%  &    7.32\%  &    6.76\% & 10.78\% \\
& Noise error              &   20.17\%  &   11.57\%  &    9.54\%  &    7.35\%  &    6.79\% & 10.89\% \\
& PGD error                &   34.13\%  &   17.05\%  &   13.40\%  &   10.37\%  &    8.98\% & 16.40\% \\
& MILP bound               &   34.52\%  &   17.12\%  &   13.46\%  &   10.41\%  &    9.05\% & {\bf 16.51\%} \\
& Dual bound               &   35.09\%  &   17.37\%  &   13.86\%  &   10.66\%  &    9.22\% & 16.84\% \\
\midrule
\multirow{5}*{Targeted robust training}
& Relative error           &   14.33\%  &    7.31\%  &    5.71\%  &    {\bf 3.24\%}  &    {\bf 2.78\%} &  6.53\% \\
& Noise error              &   20.68\%  &    8.76\%  &    6.31\%  &    3.64\%  &    3.03\% &  8.21\% \\
& PGD error                &  102.70\%  &   35.85\%  &   20.41\%  &   11.76\%  &    8.92\% & 34.26\% \\
& MILP bound               &  107.48\%  &   37.16\%  &   20.91\%  &   {\bf 11.97\%}  &    {\bf 9.03\%} & 35.55\% \\
& Dual bound               &  118.42\%  &   40.84\%  &   22.43\%  &   12.59\%  &    9.34\% & 38.78\% \\
\bottomrule
\vspace{-3em}
\end{tabular}
\end{center}
\end{table*}

\section{EXPERIMENTS}

\subsection{Fuel injection data}
We begin with a brief description of the fuel injection dataset used in this work. 
%Fuel injection data was collected as follows, using the setup depicted in \ref{figure:fuelinjection}. 
The features consist of a time sequence of $K-1$ readings of the first sensor and an additional reading of the second sensor, resulting in a feature vector of dimension $K$. The input features as well as the output are all normalized, resulting in a range from 0 to 1. We collected 20,000 data points for a training set, and 1,000 data points for a test set. 

\subsection{Architecture, parameters, and evaluation metrics}
We adopt fairly standard practices for training the neural network, which we outline here. The fuel injection virtual sensor is a neural network with a single hidden layer with 32 units and ReLU non-linearities. Note that the size of the architecture must be relatively small and simple to accomodate the engine controller. We train the network for 1000 epochs with minibatch size 512, using a stochastic gradient descent optimizer with momentum set to 0.9 using a cyclic learning rate \cite{smith2017cyclical} which peaks at 0.035 at the 250th epoch. We report mean relative error (MRE), which is computed for examples $\{x_i, y_i\}_{i=1\dots n}$ on a given neural network $f$ as 
\begin{equation}
MRE(f) = \frac{1}{n}\sum \frac{|f(x_i) - y_i|}{|y_i|}
\end{equation}
We also report the mean error under noise, where noise is drawn uniformly at random from the perturbation set and performance is averaged over 1000 draws, as well as the empirical adversarial error rate from a PGD adversary. Finally, we report exact verification results from a MILP solver and the looser duality based bound on the worst-case adversarial error rate, which constitute our provable guarantees on the performance of the controller. All experiments are compactly summarized in Table \ref{table:performance}. 

\begin{figure}[t]
      \centering
      \vspace{0.3em}
      \parbox{3in}{\centering\includegraphics[width=2.8in]{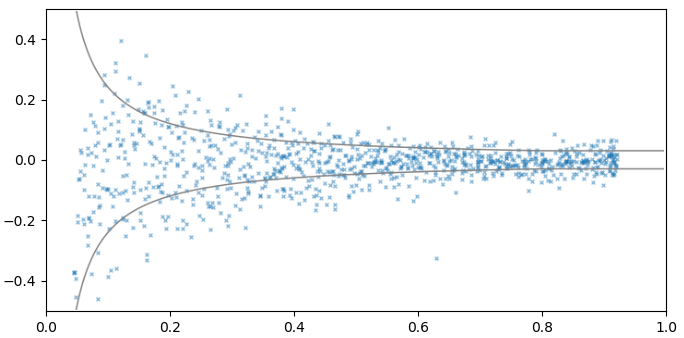}}  
      \caption{Test performance of a standard (non-robust) model over the range fuel injection quantities. Each point represents a single example, where the vertical axis denotes the relative error of the model and the horizontal axis denotes the target output quantity to be predicted as a fraction of the maximum possible value. The gray lines denote margins of allowable relative error, met by most of the data in the upper ranges.}
      \label{figure:cone}
\vspace{-1.3em}
\end{figure}

Training the fuel injection virtual sensor on this data using standard methods (minimizing mean squared error) achieves a network with 6.6\% mean relative error, and under 3\% mean relative error for the fuel injection range of 0.6 to 1.0. A scatter plot depicting the test error over different fuel injection quantities is shown in Figure \ref{figure:cone}. In addition to a standard model, we also train a model with data augmentation with non-adversarial noisy examples as an additional natural baseline. Noise for each example was uniformly randomly sampled from the target perturbation set, and the resulting trained model obtains similar relative error to the standard baseline but has slightly better performance against random noise. %The performance of both baselines are presented in more detail in Table \ref{table:performance}. 

%\begin{figure}[thpb]
%      \centering
%      \framebox{\parbox{3in}{Possible picture of the fuel injection data collection setup
%}}
%      %\includegraphics[scale=1.0]{figurefile}
%      \caption{The experimental setup for collecting fuel injection data}
%      \label{figure:fuelinjection}
%\end{figure}

\subsection{Adversarial examples for fuel injection systems}
We first demonstrate the vulnerability of the neural network virtual sensor to adversarial examples when trained using standard methods. Based on the noise levels observed in the sensor measurements, we assume a noise level of 1\% for each entry in the time series of the first sensor and 0.1\% for the entry of the second sensor, which constitutes the set of allowable perturbations that we wish to be robust to, and additionally clip the perturbed features to stay within the normalized range of 0 to 1. To generate adversarial examples and find worst-case perturbations, we use a 10 step projected gradient descent (PGD) adversary with step size 0.25\% for the time series features and 0.025\% for the remaining feature.

\begin{figure}[t]
      \centering
      \parbox{3in}{\centering\includegraphics[width=2.3in]{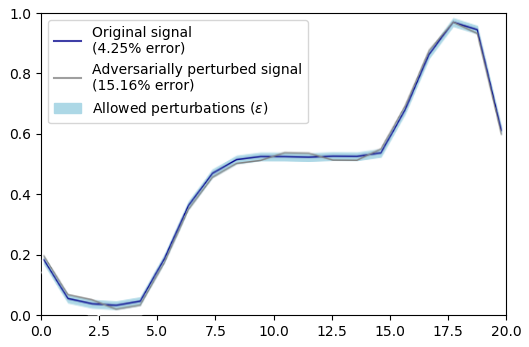}}
      \caption{An example of an adversarially perturbed time series sequence for a non-robust fuel injection controller, where the horizontal axis denotes time and the vertical axis denotes the sensor reading. The dark blue line denotes the true underlying signal, while the gray line denotes the adversarially perturbed signal. The light blue region denotes the set of allowable perturbations. This adversarial perturbation, which is well within the realm of reasonable sensor noise, increases the relative error of a standard model on this example from 4\% to 15\%}
      \label{figure:perturbation}
\vspace{-1em}
\end{figure}

We immediately find that both baselines (standard training and noisy training) are highly susceptible to adversarial examples, with the standard model having mean relative error increased from 6.61\% to 43.88\%, while the model trained on noisy examples fairs slightly better, going from 6.79\% to 33.38\% mean relative error.
% as seen in Table \ref{table:performance}. 
An example of an adversarially perturbed time series sequence is depicted in Figure \ref{figure:perturbation}. 

\subsection{Robust training and duality-base certificates}

We next train a model using provably robust training methods adapted to the regression problem by minimizing the robust mean squared error as discussed in Section \ref{section:robustregression}. As expected, the robustly trained model has the most resilience to adversarial attacks, achieving 16.40\% mean relative error under adversarial perturbations, significantly better than the baselines. Furthermore, the robust training method comes with a duality based bound which is a provable upper bound on the relative error of the neural network over the perturbation set. We find that the robustly trained network has a dual bound which guarantees at most 16.84\% worst-case mean relative error, which is a significantly better guarantee than the dual bound evaluated on the standard and noisy baselines which can only guarantee at most 50.09\% and 40.29\% worst-case mean relative error respectively. However, this comes at a cost: the performance on unperturbed examples is degraded to 10.78\%, approximately 4\% worse than the standard model. 

\subsection{Exact robustness verification}
Due to the small size of the network, it is computationally feasible calculate the exact worst case performance within the perturbation set, using an MILP solver \cite{tjeng2017evaluating}. Verifying all of the examples in the test set for a single model takes approximately 15-20 minutes. We find that the MILP bound is slightly tighter for the baseline, improving their verified performance by about 5\% mean relative error each, whereas the dual bound for the robustly trained network is already nearly tight and closely matches the MILP bound.

\begin{figure}[t]
      \centering
      \parbox{3in}{\centering\includegraphics[width=2.8in]{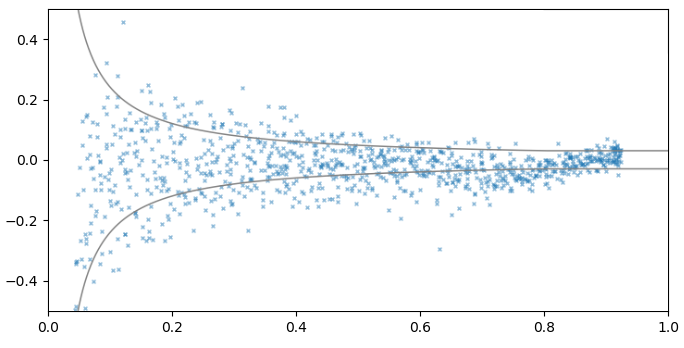}}
      
      \caption{Test performance of a model targeted to be robust over the higher range of fuel injection quantities. Each point represents a single example, where the vertical axis denotes the relative error of the model and the horizontal axis denotes the target output quantity to be predicted as a fraction of the maximum possible value. The gray lines denote margins of allowable relative error, met by most of the data in the upper ranges.}
      \label{figure:cone_robust}
\vspace{-1em}
\end{figure}

\subsection{Targeted robust training}
Finally, depending on the desired specifications of the model, it may be acceptable for the virtual sensor to perform well only on certain output ranges of fuel injection. For example, in our setting, it is typical for performance in the upper ranges (e.g. 0.6-1.0) of fuel injection to carry higher importance than the lower ranges (e.g. 0.0-0.6), and so one might wish to maximize robustness only within this range. However, training on the full range of fuel injection data is still necessary to achieve good generalization on the test set. To balance this trade-off between standard performance and robustness among the higher ranges of fuel injection, we train on a convex combination of the standard mean squared error of the entire training data and the robust mean squared error of the targeted subset of the training data. Specifically, for a dataset $(X,Y)$, we minimize the following targeted loss 
\begin{equation}
\ell_{tar}(f(X),Y) = \lambda \ell(f(X),Y) + (1-\lambda)\ell_{rob}(f(X_{tar}), Y_{tar})
\end{equation}
where $\ell$ is the standard loss (mean squared error), $\ell_{rob}$ is the robust loss (robust mean squared error), $(X_{tar}, Y_{tar})$ are the targeted subsets of the original dataset (data points with fuel injection targets from 0.6-1.0), and $\lambda$ is a hyperparameter used to control the balance between robustness and accuracy. 

Setting $\lambda = 0.8$, we are able to train a robust fuel injection virtual sensor which is specifically trained to be robust within the range 0.6-1.0. In fact, with this trade off we are able to nearly match the original standard performance of the baseline standard model with 6.53\% mean relative error on unperturbed examples. Figure \ref{figure:cone_robust} depicts the performance of the targeted robust model over various fuel injection quantities, where the model is in fact able to achieve low relative error within the range of 0.6-1.0. 

While it achieves similar relative error to the standard model, more importantly, we find that the model trained with targeted robust training achieves better verified robust performance. The guarantees obtained by the targeted robust model within the range of 0.6-1.0 are nearly on par with the model trained with pure robust training, achieving a guarantee of at most 10.69\% worst-case relative error within the perturbation set for higher injection quantities, without sacrificing standard performance. 

\section{CONCLUSION}
Although work in adversarial robustness has largely focused on large-scale vision problems, in this work we showed how provably robust methods can be leveraged for smaller, model verification problems in the regression setting to get provable specifications. Specifically, we showed how methods from duality-based approaches for training provably robust neural networks can be used to train robust virtual sensors for fuel injection with guarantees on the performance under noise. While focusing solely on robust performance results in a degradation of standard performance, if performance on only a subset of the target range is desired, then it is possible to get the best of both worlds and achieve competitive standard performance while improving robust guarantees on the target range. We hope that this work opens up more higher-stakes applications to using neural network models by demonstrating that neural networks can be trained to have reasonable provable guarantees that restore confidence in their ability to perform robustly and reliably.

\bibliographystyle{IEEEtran}
\bibliography{IEEE}

\begin{thebibliography}{10}
\providecommand{\url}[1]{#1}
\csname url@rmstyle\endcsname
\providecommand{\newblock}{\relax}
\providecommand{\bibinfo}[2]{#2}
\providecommand\BIBentrySTDinterwordspacing{\spaceskip=0pt\relax}
\providecommand\BIBentryALTinterwordstretchfactor{4}
\providecommand\BIBentryALTinterwordspacing{\spaceskip=\fontdimen2\font plus
\BIBentryALTinterwordstretchfactor\fontdimen3\font minus
  \fontdimen4\font\relax}
\providecommand\BIBforeignlanguage[2]{{%
\expandafter\ifx\csname l@#1\endcsname\relax
\typeout{** WARNING: IEEEtran.bst: No hyphenation pattern has been}%
\typeout{** loaded for the language `#1'. Using the pattern for}%
\typeout{** the default language instead.}%
\else
\language=\csname l@#1\endcsname
\fi
#2}}

\bibitem{szegedy2014intriguing}
\BIBentryALTinterwordspacing
C.~Szegedy, W.~Zaremba, I.~Sutskever, J.~Bruna, D.~Erhan, I.~Goodfellow, and
  R.~Fergus, ``Intriguing properties of neural networks,'' in
  \emph{International Conference on Learning Representations}, 2014. [Online].
  Available: \url{http://arxiv.org/abs/1312.6199}
\BIBentrySTDinterwordspacing

\bibitem{goodfellow2015explaining}
\BIBentryALTinterwordspacing
I.~Goodfellow, J.~Shlens, and C.~Szegedy, ``Explaining and harnessing
  adversarial examples,'' in \emph{International Conference on Learning
  Representations}, 2015. [Online]. Available:
  \url{http://arxiv.org/abs/1412.6572}
\BIBentrySTDinterwordspacing

\bibitem{sharif2018suitability}
M.~Sharif, L.~Bauer, and M.~K. Reiter, ``On the suitability of lp-norms for
  creating and preventing adversarial examples,'' in \emph{Proceedings of the
  IEEE Conference on Computer Vision and Pattern Recognition Workshops}, 2018,
  pp. 1605--1613.

\bibitem{tramer2019adversarial}
F.~Tram{\`e}r and D.~Boneh, ``Adversarial training and robustness for multiple
  perturbations,'' \emph{arXiv preprint arXiv:1904.13000}, 2019.

\bibitem{maini2019adversarial}
P.~Maini, E.~Wong, and J.~Z. Kolter, ``Adversarial robustness against the union
  of multiple perturbation models,'' \emph{arXiv preprint arXiv:1909.04068},
  2019.

\bibitem{engstrom2017rotation}
L.~Engstrom, B.~Tran, D.~Tsipras, L.~Schmidt, and A.~Madry, ``A rotation and a
  translation suffice: Fooling cnns with simple transformations,'' \emph{arXiv
  preprint arXiv:1712.02779}, 2017.

\bibitem{xiao2018spatially}
C.~Xiao, J.-Y. Zhu, B.~Li, W.~He, M.~Liu, and D.~Song, ``Spatially transformed
  adversarial examples,'' \emph{arXiv preprint arXiv:1801.02612}, 2018.

\bibitem{wong2019wasserstein}
E.~Wong, F.~R. Schmidt, and J.~Z. Kolter, ``Wasserstein adversarial examples
  via projected sinkhorn iterations,'' \emph{arXiv preprint arXiv:1902.07906},
  2019.

\bibitem{levine2019wasserstein}
A.~Levine and S.~Feizi, ``Wasserstein smoothing: Certified robustness against
  wasserstein adversarial attacks,'' \emph{arXiv preprint arXiv: 1910.10783},
  2019.

\bibitem{li2019adversarial}
J.~B. Li, F.~R. Schmidt, and J.~Z. Kolter, ``Adversarial camera stickers: A
  physical camera attack on deep learning classifier,'' \emph{arXiv preprint
  arXiv:1904.00759}, 2019.

\bibitem{brown2017adversarial}
T.~B. Brown, D.~Man{\'e}, A.~Roy, M.~Abadi, and J.~Gilmer, ``Adversarial
  patch,'' \emph{arXiv preprint arXiv:1712.09665}, 2017.

\bibitem{wu2019making}
Z.~Wu, S.-N. Lim, L.~Davis, and T.~Goldstein, ``Making an invisibility cloak:
  Real world adversarial attacks on object detectors,'' \emph{arXiv preprint
  arXiv:1910.14667}, 2019.

\bibitem{sharif2017adversarial}
M.~Sharif, S.~Bhagavatula, L.~Bauer, and M.~K. Reiter, ``Adversarial generative
  nets: Neural network attacks on state-of-the-art face recognition,''
  \emph{arXiv preprint arXiv:1801.00349}, 2017.

\bibitem{pmlr-v80-athalye18b}
\BIBentryALTinterwordspacing
A.~Athalye, L.~Engstrom, A.~Ilyas, and K.~Kwok, ``Synthesizing robust
  adversarial examples,'' in \emph{Proceedings of the 35th International
  Conference on Machine Learning}, ser. Proceedings of Machine Learning
  Research, J.~Dy and A.~Krause, Eds., vol.~80.\hskip 1em plus 0.5em minus
  0.4em\relax Stockholmsmässan, Stockholm Sweden: PMLR, 10--15 Jul 2018, pp.
  284--293. [Online]. Available:
  \url{http://proceedings.mlr.press/v80/athalye18b.html}
\BIBentrySTDinterwordspacing

\bibitem{madry2018towards}
\BIBentryALTinterwordspacing
A.~Madry, A.~Makelov, L.~Schmidt, D.~Tsipras, and A.~Vladu, ``Towards deep
  learning models resistant to adversarial attacks,'' in \emph{International
  Conference on Learning Representations}, 2018. [Online]. Available:
  \url{https://openreview.net/forum?id=rJzIBfZAb}
\BIBentrySTDinterwordspacing

\bibitem{zhang2019theoretically}
H.~Zhang, Y.~Yu, J.~Jiao, E.~P. Xing, L.~E. Ghaoui, and M.~I. Jordan,
  ``Theoretically principled trade-off between robustness and accuracy,''
  \emph{arXiv preprint arXiv:1901.08573}, 2019.

\bibitem{wong2018provable}
E.~Wong and Z.~Kolter, ``Provable defenses against adversarial examples via the
  convex outer adversarial polytope,'' in \emph{International Conference on
  Machine Learning}, 2018, pp. 5283--5292.

\bibitem{wong2018scaling}
\BIBentryALTinterwordspacing
E.~Wong, F.~Schmidt, J.~H. Metzen, and J.~Z. Kolter, ``Scaling provable
  adversarial defenses,'' in \emph{Advances in Neural Information Processing
  Systems 31}, S.~Bengio, H.~Wallach, H.~Larochelle, K.~Grauman,
  N.~Cesa-Bianchi, and R.~Garnett, Eds.\hskip 1em plus 0.5em minus 0.4em\relax
  Curran Associates, Inc., 2018, pp. 8410--8419. [Online]. Available:
  \url{http://papers.nips.cc/paper/8060-scaling-provable-adversarial-defenses.pdf}
\BIBentrySTDinterwordspacing

\bibitem{raghunathan2018semi}
\BIBentryALTinterwordspacing
A.~Raghunathan, J.~Steinhardt, and P.~S. Liang, ``Semidefinite relaxations for
  certifying robustness to adversarial examples,'' in \emph{Advances in Neural
  Information Processing Systems 31}, S.~Bengio, H.~Wallach, H.~Larochelle,
  K.~Grauman, N.~Cesa-Bianchi, and R.~Garnett, Eds.\hskip 1em plus 0.5em minus
  0.4em\relax Curran Associates, Inc., 2018, pp. 10\,900--10\,910. [Online].
  Available:
  \url{http://papers.nips.cc/paper/8285-semidefinite-relaxations-for-certifying-robustness-to-adversarial-examples.pdf}
\BIBentrySTDinterwordspacing

\bibitem{mirman2018diff}
\BIBentryALTinterwordspacing
M.~Mirman, T.~Gehr, and M.~Vechev, ``Differentiable abstract interpretation for
  provably robust neural networks,'' in \emph{International Conference on
  Machine Learning (ICML)}, 2018. [Online]. Available:
  \url{https://www.icml.cc/Conferences/2018/Schedule?showEvent=2477}
\BIBentrySTDinterwordspacing

\bibitem{gowal2018interval}
\BIBentryALTinterwordspacing
S.~Gowal, K.~Dvijotham, R.~Stanforth, R.~Bunel, C.~Qin, J.~Uesato,
  R.~Arandjelovic, T.~A. Mann, and P.~Kohli, ``On the effectiveness of interval
  bound propagation for training verifiably robust models,'' \emph{CoRR}, vol.
  abs/1810.12715, 2018. [Online]. Available:
  \url{http://arxiv.org/abs/1810.12715}
\BIBentrySTDinterwordspacing

\bibitem{jia2019certified}
R.~Jia, A.~Raghunathan, K.~G{\"o}ksel, and P.~Liang, ``Certified robustness to
  adversarial word substitutions,'' \emph{arXiv preprint arXiv:1909.00986},
  2019.

\bibitem{tramer2017ensemble}
F.~Tram{\`e}r, A.~Kurakin, N.~Papernot, I.~Goodfellow, D.~Boneh, and
  P.~McDaniel, ``Ensemble adversarial training: Attacks and defenses,''
  \emph{arXiv preprint arXiv:1705.07204}, 2017.

\bibitem{kurakin2017adversarial}
\BIBentryALTinterwordspacing
A.~Kurakin, I.~Goodfellow, and S.~Bengio, ``Adversarial examples in the
  physical world,'' \emph{ICLR Workshop}, 2017. [Online]. Available:
  \url{https://arxiv.org/abs/1607.02533}
\BIBentrySTDinterwordspacing

\bibitem{dong2018boosting}
Y.~Dong, F.~Liao, T.~Pang, H.~Su, J.~Zhu, X.~Hu, and J.~Li, ``Boosting
  adversarial attacks with momentum,'' in \emph{Proceedings of the IEEE
  conference on computer vision and pattern recognition}, 2018, pp. 9185--9193.

\bibitem{mosbach2018logit}
M.~Mosbach, M.~Andriushchenko, T.~Trost, M.~Hein, and D.~Klakow, ``Logit
  pairing methods can fool gradient-based attacks,'' \emph{arXiv preprint
  arXiv:1810.12042}, 2018.

\bibitem{yang2019me}
Y.~Yang, G.~Zhang, D.~Katabi, and Z.~Xu, ``Me-net: Towards effective
  adversarial robustness with matrix estimation,'' \emph{arXiv preprint
  arXiv:1905.11971}, 2019.

\bibitem{shafahi2019adversarial}
A.~Shafahi, M.~Najibi, A.~Ghiasi, Z.~Xu, J.~Dickerson, C.~Studer, L.~S. Davis,
  G.~Taylor, and T.~Goldstein, ``Adversarial training for free!'' \emph{arXiv
  preprint arXiv:1904.12843}, 2019.

\bibitem{zhang2019you}
D.~Zhang, T.~Zhang, Y.~Lu, Z.~Zhu, and B.~Dong, ``You only propagate once:
  Painless adversarial training using maximal principle,'' \emph{arXiv preprint
  arXiv:1905.00877}, 2019.

\bibitem{wong2019fast}
E.~Wong, L.~Rice, and J.~Z. Kolter, ``Fast is better than free: Revisiting
  adversarial training,'' in \emph{International Conference on Learning
  Representations}, 2019.

\bibitem{guo2017countering}
C.~Guo, M.~Rana, M.~Cisse, and L.~Van Der~Maaten, ``Countering adversarial
  images using input transformations,'' \emph{arXiv preprint arXiv:1711.00117},
  2017.

\bibitem{buckman2018thermometer}
J.~Buckman, A.~Roy, C.~Raffel, and I.~Goodfellow, ``Thermometer encoding: One
  hot way to resist adversarial examples,'' 2018.

\bibitem{song2017pixeldefend}
Y.~Song, T.~Kim, S.~Nowozin, S.~Ermon, and N.~Kushman, ``Pixeldefend:
  Leveraging generative models to understand and defend against adversarial
  examples,'' \emph{arXiv preprint arXiv:1710.10766}, 2017.

\bibitem{metzen2017detecting}
J.~H. Metzen, T.~Genewein, V.~Fischer, and B.~Bischoff, ``On detecting
  adversarial perturbations,'' \emph{arXiv preprint arXiv:1702.04267}, 2017.

\bibitem{feinman2017detecting}
R.~Feinman, R.~R. Curtin, S.~Shintre, and A.~B. Gardner, ``Detecting
  adversarial samples from artifacts,'' \emph{arXiv preprint arXiv:1703.00410},
  2017.

\bibitem{papernot2016distillation}
N.~Papernot, P.~McDaniel, X.~Wu, S.~Jha, and A.~Swami, ``Distillation as a
  defense to adversarial perturbations against deep neural networks,'' in
  \emph{Security and Privacy (SP), 2016 IEEE Symposium on}.\hskip 1em plus
  0.5em minus 0.4em\relax IEEE, 2016, pp. 582--597.

\bibitem{lu2017no}
J.~Lu, H.~Sibai, E.~Fabry, and D.~Forsyth, ``No need to worry about adversarial
  examples in object detection in autonomous vehicles,'' \emph{arXiv preprint
  arXiv:1707.03501}, 2017.

\bibitem{kannan2018adversarial}
H.~Kannan, A.~Kurakin, and I.~Goodfellow, ``Adversarial logit pairing,''
  \emph{arXiv preprint arXiv:1803.06373}, 2018.

\bibitem{tao2018attacks}
G.~Tao, S.~Ma, Y.~Liu, and X.~Zhang, ``Attacks meet interpretability:
  Attribute-steered detection of adversarial samples,'' in \emph{Advances in
  Neural Information Processing Systems}, 2018, pp. 7717--7728.

\bibitem{carlini2016defensive}
N.~Carlini and D.~Wagner, ``Defensive distillation is not robust to adversarial
  examples,'' \emph{arXiv preprint arXiv:1607.04311}, 2016.

\bibitem{carlini2017towards}
------, ``Towards evaluating the robustness of neural networks,'' in
  \emph{Security and Privacy (SP), 2017 IEEE Symposium on}.\hskip 1em plus
  0.5em minus 0.4em\relax IEEE, 2017, pp. 39--57.

\bibitem{athalye2017synthesizing}
A.~Athalye, L.~Engstrom, A.~Ilyas, and K.~Kwok, ``Synthesizing robust
  adversarial examples,'' \emph{arXiv preprint arXiv:1707.07397}, 2017.

\bibitem{engstrom2018evaluating}
L.~Engstrom, A.~Ilyas, and A.~Athalye, ``Evaluating and understanding the
  robustness of adversarial logit pairing,'' \emph{arXiv preprint
  arXiv:1807.10272}, 2018.

\bibitem{carlini2019ami}
N.~Carlini, ``Is ami (attacks meet interpretability) robust to adversarial
  examples?'' \emph{arXiv preprint arXiv:1902.02322}, 2019.

\bibitem{uesato2018adversarial}
J.~Uesato, B.~O'Donoghue, A.~v.~d. Oord, and P.~Kohli, ``Adversarial risk and
  the dangers of evaluating against weak attacks,'' \emph{arXiv preprint
  arXiv:1802.05666}, 2018.

\bibitem{athalye2018obfuscated}
A.~Athalye, N.~Carlini, and D.~Wagner, ``Obfuscated gradients give a false
  sense of security: Circumventing defenses to adversarial examples,''
  \emph{arXiv preprint arXiv:1802.00420}, 2018.

\bibitem{carlini2019evaluating}
N.~Carlini, A.~Athalye, N.~Papernot, W.~Brendel, J.~Rauber, D.~Tsipras,
  I.~Goodfellow, and A.~Madry, ``On evaluating adversarial robustness,''
  \emph{arXiv preprint arXiv:1902.06705}, 2019.

\bibitem{katz2017reluplex}
G.~Katz, C.~Barrett, D.~L. Dill, K.~Julian, and M.~J. Kochenderfer, ``Reluplex:
  An efficient smt solver for verifying deep neural networks,'' in
  \emph{International Conference on Computer Aided Verification}.\hskip 1em
  plus 0.5em minus 0.4em\relax Springer, 2017, pp. 97--117.

\bibitem{huang2017safety}
X.~Huang, M.~Kwiatkowska, S.~Wang, and M.~Wu, ``Safety verification of deep
  neural networks,'' in \emph{International Conference on Computer Aided
  Verification}.\hskip 1em plus 0.5em minus 0.4em\relax Springer, 2017, pp.
  3--29.

\bibitem{ehlers2017formal}
R.~Ehlers, ``Formal verification of piece-wise linear feed-forward neural
  networks,'' in \emph{International Symposium on Automated Technology for
  Verification and Analysis}.\hskip 1em plus 0.5em minus 0.4em\relax Springer,
  2017, pp. 269--286.

\bibitem{tjeng2017evaluating}
V.~Tjeng, K.~Xiao, and R.~Tedrake, ``Evaluating robustness of neural networks
  with mixed integer programming,'' \emph{arXiv preprint arXiv:1711.07356},
  2017.

\bibitem{raghunathan2018semidefinite}
A.~Raghunathan, J.~Steinhardt, and P.~S. Liang, ``Semidefinite relaxations for
  certifying robustness to adversarial examples,'' in \emph{Advances in Neural
  Information Processing Systems}, 2018, pp. 10\,877--10\,887.

\bibitem{salman2019convex}
H.~Salman, G.~Yang, H.~Zhang, C.-J. Hsieh, and P.~Zhang, ``A convex relaxation
  barrier to tight robust verification of neural networks,'' \emph{arXiv
  preprint arXiv:1902.08722}, 2019.

\bibitem{sinha2017certifying}
A.~Sinha, H.~Namkoong, and J.~Duchi, ``Certifying some distributional
  robustness with principled adversarial training,'' \emph{arXiv preprint
  arXiv:1710.10571}, 2017.

\bibitem{raghunathan2018certified}
A.~Raghunathan, J.~Steinhardt, and P.~Liang, ``Certified defenses against
  adversarial examples,'' \emph{arXiv preprint arXiv:1801.09344}, 2018.

\bibitem{gehr2018ai2}
T.~Gehr, M.~Mirman, D.~Drachsler-Cohen, P.~Tsankov, S.~Chaudhuri, and
  M.~Vechev, ``Ai2: Safety and robustness certification of neural networks with
  abstract interpretation,'' in \emph{2018 IEEE Symposium on Security and
  Privacy (SP)}.\hskip 1em plus 0.5em minus 0.4em\relax IEEE, 2018, pp. 3--18.

\bibitem{mirman2018differentiable}
M.~Mirman, T.~Gehr, and M.~Vechev, ``Differentiable abstract interpretation for
  provably robust neural networks,'' in \emph{International Conference on
  Machine Learning}, 2018, pp. 3575--3583.

\bibitem{gowal2018effectiveness}
S.~Gowal, K.~Dvijotham, R.~Stanforth, R.~Bunel, C.~Qin, J.~Uesato, T.~Mann, and
  P.~Kohli, ``On the effectiveness of interval bound propagation for training
  verifiably robust models,'' \emph{arXiv preprint arXiv:1810.12715}, 2018.

\bibitem{cohen2019certified}
J.~M. Cohen, E.~Rosenfeld, and J.~Z. Kolter, ``Certified adversarial robustness
  via randomized smoothing,'' \emph{arXiv preprint arXiv:1902.02918}, 2019.

\bibitem{salman2019provably}
H.~Salman, G.~Yang, J.~Li, P.~Zhang, H.~Zhang, I.~Razenshteyn, and S.~Bubeck,
  ``Provably robust deep learning via adversarially trained smoothed
  classifiers,'' \emph{arXiv preprint arXiv:1906.04584}, 2019.

\bibitem{xiao2018training}
K.~Y. Xiao, V.~Tjeng, N.~M. Shafiullah, and A.~Madry, ``Training for faster
  adversarial robustness verification via inducing relu stability,''
  \emph{arXiv preprint arXiv:1809.03008}, 2018.

\bibitem{croce2018provable}
\BIBentryALTinterwordspacing
F.~Croce, M.~Andriushchenko, and M.~Hein, ``Provable robustness of relu
  networks via maximization of linear regions,'' \emph{CoRR}, vol.
  abs/1810.07481, 2018. [Online]. Available:
  \url{http://arxiv.org/abs/1810.07481}
\BIBentrySTDinterwordspacing

\bibitem{lygeros1998verified}
J.~Lygeros, D.~N. Godbole, and S.~Sastry, ``Verified hybrid controllers for
  automated vehicles,'' \emph{IEEE transactions on automatic control}, vol.~43,
  no.~4, pp. 522--539, 1998.

\bibitem{gerber2010complete}
C.~Gerber, S.~Preu{\ss}e, and H.-M. Hanisch, ``A complete framework for
  controller verification in manufacturing,'' in \emph{2010 IEEE 15th
  Conference on Emerging Technologies \& Factory Automation (ETFA 2010)}.\hskip
  1em plus 0.5em minus 0.4em\relax IEEE, 2010, pp. 1--9.

\bibitem{schumann2003verification}
J.~Schumann, P.~Gupta, and S.~Nelson, ``On verification \& validation of neural
  network based controllers,'' in \emph{Proc. of International Conf on
  Engineering Applications of Neural Networks, EANN}, vol.~3, 2003, pp.
  401--47.

\bibitem{gupta2004tool}
P.~Gupta and J.~Schumann, ``A tool for verification and validation of neural
  network based adaptive controllers for high assurance systems,'' 2004.

\bibitem{scheibler2015towards}
K.~Scheibler, L.~Winterer, R.~Wimmer, and B.~Becker, ``Towards verification of
  artificial neural networks.'' in \emph{MBMV}, 2015, pp. 30--40.

\bibitem{sun2019formal}
X.~Sun, H.~Khedr, and Y.~Shoukry, ``Formal verification of neural network
  controlled autonomous systems,'' in \emph{Proceedings of the 22nd ACM
  International Conference on Hybrid Systems: Computation and Control}.\hskip
  1em plus 0.5em minus 0.4em\relax ACM, 2019, pp. 147--156.

\bibitem{smith2017cyclical}
L.~N. Smith, ``Cyclical learning rates for training neural networks,'' in
  \emph{2017 IEEE Winter Conference on Applications of Computer Vision
  (WACV)}.\hskip 1em plus 0.5em minus 0.4em\relax IEEE, 2017, pp. 464--472.

\end{thebibliography}

\end{document}